\documentclass[preprint,12pt]{elsarticle}

\usepackage{amssymb}
\journal{Journal}
\usepackage{xcolor}
\usepackage{multirow}
\usepackage{float}
\usepackage{hyperref}
\usepackage{comment}
\usepackage{algorithm}
\usepackage{algpseudocode}
\usepackage{amsmath}

\usepackage[nameinlink]{cleveref} % Para referencias cruzadas inteligentes

\begin{document}

\begin{frontmatter}

\title{Continual Learning at the Edge: An Agnostic IIoT Architecture}

\author{Pablo García-Santaclara}
\cortext[pgs]{Corresponding author}
\ead{pgarcia@alumnos.uvigo.es}

\author{Bruno Fernández-Castro}
\ead{bfernandez@gradiant.org}

\author{Rebeca P. Díaz-Redondo}
\ead{rebeca@det.uvigo.es}

\author{Carlos Calvo-Moa}
\ead{ccalvo@gradiant.org}

\author{Henar Mariño-Bodelón}
\ead{hmarino@gradiant.org}

\address{atlanTTic - I\&C Lab - Universidade de Vigo, Escola de Enxeñaría de Telecomunicación Campus Univesitario, Vigo 36310, Spain}

\address{Centro Tecnolóxico de Telecomunicacións de Galicia (GRADIANT), Carretera do Vilar, 56-58, Vigo 36214, Spain}

\begin{abstract}
The exponential growth of Internet-connected devices has presented challenges to traditional centralized computing systems due to latency and bandwidth limitations. Edge computing has evolved to address these difficulties by bringing computations closer to the data source. Additionally, traditional machine learning algorithms are not suitable for edge-computing systems, where data usually arrives in a dynamic and continual way. However, incremental learning offers a good solution for these settings. We introduce a new approach that applies the incremental learning philosophy within an edge-computing scenario for the industrial sector with a specific purpose: real time quality control in a manufacturing system. Applying continual learning we reduce the impact of catastrophic forgetting and provide an efficient and effective solution.  

\end{abstract}


\end{frontmatter}

\newpage

\section{Introduction}

The huge number of Internet-connected devices has led to an exponential increase in the amount of data being generated. This rapid growth does not fit well with traditional centralized computing architectures. Cloud computing~\cite{cloud_computing}, while providing scalable and centralized data processing capabilities, faces challenges that limit its effectiveness in certain scenarios. One of the main problems is the latency introduced by transmitting large volumes of data to distant centralized data centers. Other problem is that cloud infrastructures tend to be inefficient in terms of computation, memory and storage, consuming significant amounts of energy. Within this scenario, edge computing~\cite{edge_overview,edge_promise} emerges as a solution to address the limitations of traditional cloud computing, particularly in managing the increasing volume of data generated by edge devices. Edge computing allows data to be processed in low-powered devices closer to the sources, reducing latency, and improving security by keeping data local and optimizing bandwidth and power usage. The industry is not unaware of this emerging computing model. IIoT (Industrial Internet of Things) devices and machine learning algorithms are, therefore, the key enablers of advanced edge computing systems for industry. 

 In the conventional machine learning approach, the model is trained offline with all the data at once and, when re-training is necessary due to a decrease in accuracy, a complete re-training is performed with a larger dataset. However, this is not the best solution considering the efficiency of computing resources for borderline industrial scenarios where non-stationary data from manufacturing processes are constantly arriving. Incremental Learning~\cite{incremental,incremental2} is a machine learning paradigm in which an AI model learns incrementally, new information is acquired continuously and integrated with existing knowledge. One of the main challenges of incremental learning is the difficulty of learning new tasks without forgetting previously acquired knowledge. This phenomenon, known as catastrophic forgetting~\cite{catastrophic_forgetting_in_connectionist_networks,Measuring_Catastrophic_forgetting_in_NN,Overcoming_catastrophic_forgetting}, occurs because neural network weights that were important for previous tasks are modified to suit the purposes of new tasks. Lifelong Learning~\cite{lifelong_machine_learning, Continual_lifelong_learning_nn} is a machine learning field focused on designing systems capable of learning and retaining knowledge over time. The accumulated knowledge is not only useful for current tasks but can also be used to improve at new ones. The integration of new information should improve the accuracy of related prior knowledge.

 Thus, lifelong learning methods are promising for real-world edge computing industrial applications and autonomous agents (i.e. real-time quality control, defeat detection, fault diagnosis, etc.) as they are able to efficiently update a machine learning model from new information requiring less compute, rather than continuously retraining from scratch. 

Within this context, we propose a solution based on the continual learning philosophy to be applied in the industrial sector over an edge-computing infrastructure. We identify two main contributions. First, we define an agnostic edge computing architecture that allows the real-time processing of non-stationary data streams through continual learning models in manufacturing environments. Since it does not depend on specific kind of data or edge computing devices or even industrial processes, it may be applied in many different scenarios. Second, we have implemented this solution on a real industrial setting. With this aim, we have taken advantage of the resources of the AI-REDGIO 5.0~\cite{airedgio} project (funded by the European Commission) to find an appropriate use case. Since this project focuses on the implementation of competitive AI-at-the-Edge Digital Transformation of Industry 5.0 Manufacturing Small and Medium Enterprises, we have selected a cheese producing SME, Quescrem, located in the north of Spain. Specifically, we have focused on its quality control process. We have validated our proposal in this scenario with promising results. 

The paper is organized as follows. \Cref{relatedwork} summarizes the advances in the edge-computing area and overviews different approaches for continual learning within this field. \Cref{proposal} describes the generic architecture we propose, applicable to other industrial settings. \Cref{results} details the implementation of the architecture for our use case, quality control in a cheese factory, and analyzes the obtained results. Finally, \Cref{conclusions} summarizes the main conclusions of this research work and the future directions that it opens.

\section{Related work}
\label{relatedwork}

Edge computing has shown good performance in many scenarios including 5g networks, predictive maintenance, security monitoring, video analytics, or data caching~\cite{cao2020overview,khan2019edge}. The increase in its popularity is based on numerous advantages such as: reduced latency by processing data locally, enabling devices to communicate within a network swiftly without depending on a central server. There is an increasing number of scenarios where a smaller latency is required~\cite{ferrari2017evaluation} and latency-critical IoT applications are one of the key objectives of 5G~\cite{schulz2017latency}; lower bandwidth consumption, as traditional cloud computing typically involves sending large volumes of raw data from devices to centralized servers, whereas in edge approaches the processing is done locally and only important information is transmitted; improved privacy, as local processing and the avoidance of sending all data provide a much more secure environment that is less prone to attacks; and real-time processing capability, particularly important when the analysis and response must be immediate, a common scenario in Industry 4.0 applications~\cite{7488250,8030322}. In recent years the study of deployments in edge environments has received considerable attention~\cite{cui2021survey, 8466364} with a focus on the security of the deployment, the agility of the network, and the energy consumption of the edge nodes. In \cite{li2020read} and \cite{chen2020optimal} the authors studied how to distribute application instances across edge servers within a region to maximize their collective robustness for all the users. \cite{jiang2020edge} introduces a novel approach to deploying edge computing nodes in smart manufacturing environments, a clustering algorithm that uses different factors, such as system network delay and deployment cost, to optimize the deployment. \cite{illa2018practical} presents a manual for adapting traditional manufacturing units in a smart factory, following the principles of Industry 4.0, introducing IoT and Edge technologies, and concluding that although the transformation may be costly, it ends up reducing the cost per unit. Despite the search for reduced latency, improved robustness and real-time processing capabilities, these works do not incorporate any machine learning models, and, therefore, do not focus on the complexities associated with their deployment in edge environments. Our contribution, instead, emphasizes the development of a model-agnostic architecture specifically designed to support the integration of machine learning models at the edge. The architecture we define is not tied to a specific model, allowing for greater flexibility in selecting or changing models based on specific use cases or evolving requirements. This aspect is extremely important as it concerns several key factors of the edge computing paradigm, as it was mentioned above.

If a machine learning model is able to reliably and incrementally update at the edge from the incoming data stream, some issues linked to offline training, such as non-scalable storage systems or increased computational power and costs for continuous re-training over time, disappear. Real-time on-device training through continual learning is therefore a remarkable feature that can lead to more efficient and cost-effective edge AI solutions.  However, a limited amount of works explore the use of continual learning in edge computing scenarios. Among them, a framework, known as SParCL~\cite{NEURIPS2022_80133d0f}, has been proposed, which seeks to improve the efficiency of continual learning for its use in low-resources devices. They propose different methods with the objective of learning a sparse network during the entire continual learning process, removing less informative training data and sparse gradient updates. In~\cite{demosthenous2021continuallearningedgetensorflow} the authors propose a series of modifications to TensorFlow Lite, (an adaptation of the classic machine learning library TensorFlow~\cite{abadi2016tensorflow} to run models on resource-constrained devices) to include continual learning methods, demonstrating its correct performance in edge devices on the CORe50 benchmark~\cite{lomonaco2017core50}. In~\cite{9708956} different experiments are conducted with several existing continual learning schemes and different datasets taken from mobile and embedded sensing applications, finding various conclusions about the most appropriate methods according to specific contexts.

%%%%%%%%%%%%%%%%%%%%%%%%%%%%%%%%%%%%%%%%%%%%%%%%%%%%%%%%%%%%%%%%%%%%

\section{A Continual learning architecture for IIoT}
\label{proposal}

\begin{figure}[ht]
  \centering
\includegraphics[width=\textwidth]{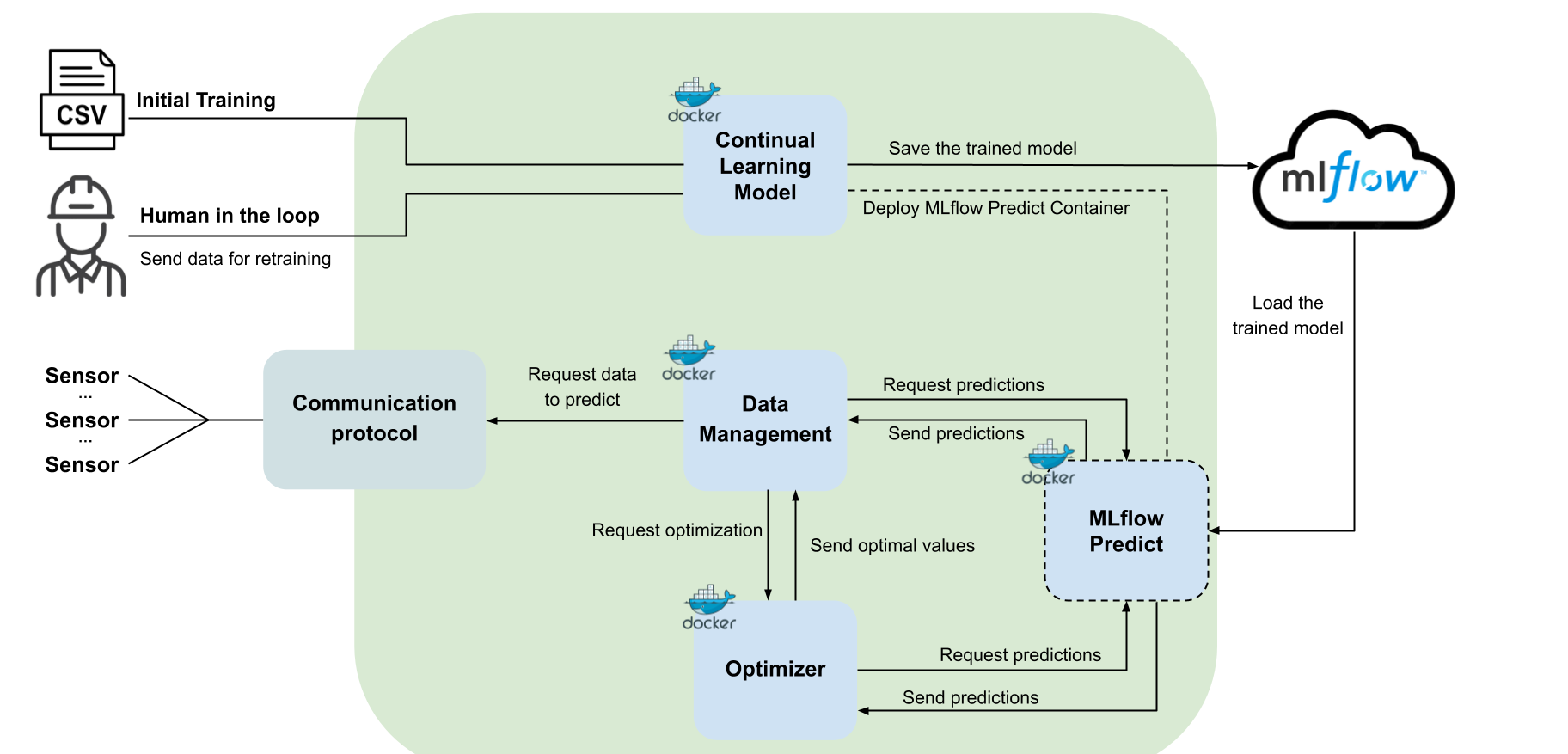}
  \caption{Edge computing architecture proposal}
  \label{fig:deployment}
\end{figure}

We propose an agnostic architecture that consists of four modules as it is shown in \autoref{fig:deployment}: data management, MLflow, model, and optimization.  

In order to deploy these four modules, we suggest using two well-known software tools: Docker~\cite{docker}, intensively used within the software production deployment field; and MLflow~\cite{MLflow}, commonly applied within the machine learning field. 

Each of the modules, depicted in \autoref{fig:deployment} are following described.

\vspace{0.3cm}

\subsection{Data Management}

The Data Management component preprocesses and manages data. It collects real-time data and generates samples that are sent to the MLflow Predict component to obtain quality predictions of the products manufactured. In case there are a relevant number of negative predictions, the data management will access the Optimizer component. Once the continual learning model is deployed, this element starts receiving data collected by sensors from different machines during the production process. For the transfer of data between the sensors in the factory and the device where the prototype is deployed, a communication protocol that transmits the data from one point to the other in an efficient and secure way is required.

\subsection{MLflow Predict}

Once the training or retraining process is completed and the continual learning model is fully trained, the MLflow Predict component can be deployed. In case a old version of this model is already deployed, the MLflow Predict container is stopped and then relaunched with the newly trained model. This element creates a server to receive samples and perform predictions using the specified model. MLflow is used to monitor the model, launch it in an automated way, and keep a record of the metrics (F1, Accuracy, etc.) and parameters of each execution carried out. 

\subsection{Continual Learning model}

It is responsible for training and evaluating the model, registering the trained model in MLflow, and deploying the MLflow prediction container. The model container receives data from two different sources. At first, pre-deployment initial training can be performed from a file with a small set of previously collected data, so that the model does not start from scratch. Afterwards, the model starts to be retrained in an online way when necessary. For this retraining, we propose the paradigm of the Human in the Loop~\cite{human_in_the_loop}, which is based on introducing the human into the retraining stream. The human in concern is required to introduce the label into the sample. Thus, in this operation, new labeled data are generated and used to retrain the model incrementally. In addition, it allows the performance of the model to be evaluated, since the prediction generated can be compared with the actual output. Once the model has been trained or retrained, the results of the model evaluations are saved using MLflow, where the model, after being serialized, is also recorded. On the other hand, the MLflow container is deployed. The model container also has its own endpoint for making predictions, instead of using MLFlow predict container for scenarios where no cloud infrastructure is available. This allows for a complete edge computing architecture, without any components or communication with the cloud. In such a case the monitoring of the model can be carried out locally in the device. The MLflow container remains essential since its function is not only to save the results, it provides an interface that makes the use of the system more accessible to users, as well as facilities such as making graphs or easily sorting the results.

\subsection{Optimizer}

This component receives the sample that produced a negative prediction (i.e., output class labeled as "1"), which means problems in the behavior of manufacturing processes, such as quality defects or abnormal deviations, and initiates a grid search based optimization process to automatically find the appropriate combination of input covariates that produce the best performing model output. This new combination will correct the fabrication problem detected by the model.  

%%%%%%%%%%%%%%%%%%%%%%%%%%%%%%%%%%%%%%%%%%
\section{Implementation and validation in a real use case}
\label{results}

After having designed the global architecture, we have also implemented and validated our proposal with a real use case. 

As it was previously mentioned, we have selected the quality control of a cheese factory (Quescrem). For this kind of manufacturing company is very important the early detection (real time) of defects in production to avoid economic costs and also the potential impact on consumer loyalty and brand reputation. 

During production, the product analysis checks the hardness of manufactured cheese, which is the key performance indicator (KPI) to be optimized together with the minimization of the amount of waste to be disposed of. Currently, defective products are discarded in the factory once the production process is finished, through laboratory analysis of the final product. With the implementation of this prototype, it is expected to be able to anticipate the final quality of the products before the laboratory analysis. The main aim of our implementation is the real-time prediction of the cheese hardness from process variables mentioned above. Once significant deviations from valid quality thresholds are detected, the optimizer will indicate correction actions in input production parameters, improving the hardness of the final product and reducing the amount of waste.

\subsection{Use case description: data and manufacturing process}
\label{data_desc}

\begin{figure}[H]
  \centering
\includegraphics[width=\textwidth]{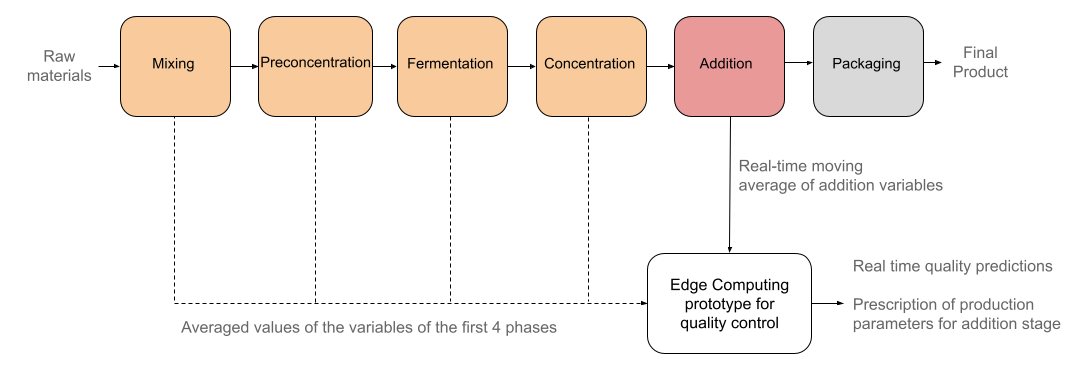}
  \caption{Serial multi-stage continuous manufacturing process scheme}
  \label{fig:scheme}
\end{figure}

The continuous manufacturing process in the Quescrem's cheese factory (depicted in \autoref{fig:scheme}) consists of six consecutive stages: mixing, pre-concentration, fermentation, concentration, addition and packaging. The production process starts by mixing initial ingredients (milk, buttermilk, etc.) until a homogeneous product is obtained. This mixture is preconcentrated and pasteurized to eliminate pathogens. Then, specific ferments are added for fermentation, allowing the mixture to rest and fermentation starts. After sufficient fermentation, the product is concentrated again. Then, in the addition phase, curds from the previous process and other ingredients are added to the mixture, resulting in the final product. This final product finally goes through another pasteurization process and is packaged.  

The penultimate stage, addition, is the one in which operators can actively make changes and try to improve the final product. The dataset for the model includes static data from the first three phases—representing the average values of variables in those phases—and dynamic data from the fifth phase, which arrives at the model periodically. It is during this stage that quality predictions for the cheese are required. %\autoref{fig:scheme} overviews this manufacturing process.

Variables collected through sensors during these six phases include temperature, pressure, pH, hardness, fat ratio, protein ratio, lactose, flow rate, viscosity, and frequency. All these data are of two types:  on the one hand, samples that are analyzed in the laboratory, and therefore there is only one record for each production batch; and samples that come from the sensors monitoring the six-staged production process, arriving with a high frequency (i.e. two samples per minute). 

\subsection{Selection of the low-powered edge device}

The edge device finally chosen, where the different modules will be depoloyed, was a Raspberry Pi 4B. The Raspberry Pi 4B device has a Broadcom BCM2711, Quad core Cortex-A72 (ARM v8) 64-bit SoC 1.8GHz
and 4GB of SDRAM, providing sufficient memory to handle the data processing requirements, and an ARM64 architecture.  The choice of the Raspberry Pi 4B was influenced by its balance between computing power, energy efficiency, and affordability, making it suitable for deployment in a manufacturing environment.

\subsection{Continual learning model for edge computing}
The continual learning model chosen was TRIL3 \cite{garciasantaclara2024overcomingcatastrophicforgettingtabular}, a framework for solving the problem of catastrophic forgetting in tabular data classification. It employs XuILVQ~\cite{gonzalez2022xuilvq}, an incremental generative model that provides synthetic data representative of past tasks, to maintain acquired knowledge without storing old samples, and Deep Neural Decision Forest (DNDF)~\cite{kontschieder2015deep}, which is a modified classification algorithm that runs incrementally. DNDF learns new classification tasks from synthetic data supplied by XuILVQ, maintaining consistent performance over time. Overall, TRIL3 allows for continual learning and adaptability to new tasks while preserving performance on past tasks, making it suited for changing industrial environments.

\subsection{Use case deployment}
Although the system could not be implemented directly in the cheese manufacturing plant at this stage of the project, we have implemented a very accurate simulation of the final scenario using all the final software tools, the edge device and real data provided by the Quescrem's cheese factory. Thus, this simulation allowed testing of the implementation using datasets created from factory data, which was instrumental in validating the continual learning model under realistic conditions. 

This validation architecture is based on the one depicted in \autoref{fig:deployment} and consists of four Docker images (one for each component). These images were uploaded to Harbor~\cite{harbor}, facilitating their deployment. The implementation also incorporates two services: MinIO~\cite{MinIO}, a cloud object store to save the models and other optional files, also called artifacts, and a PostgreSQL~\cite{PostgreSQL} cloud database, an open source database system that provides support for different SQL functions, in order to save the metrics and parameters of the models. All files that are not metrics or parameters, such as trained models, data, images or any other files to be recorded, are also stored in MinIO.
The system was deployed on the Raspberry board and the entire process was simulated using real data extracted from the factory. 

The sensors communicate with our system using Open Platform Communications Unified Architecture (OPC-UA)~\cite{opc-ua}, a standard communication protocol designed for industrial automation. We simulated the arrival of the data at the OPC-UA client using a stream of the data reserved for prediction, as the data does not arrive directly from the OPC-UA server.

When the continual learning model predicts the quality within the specified satisfactory range, the system continues to operate normally. When there is a significant number of adverse predictions (i.e. the model detects a problem with the hardness of the cheese), the optimizer module is accessed. The optimizer iterates predictions where the values of five additive process variables are changed over a bounded range of possible values in an attempt to find the set of values that maximizes the probability of good final quality. 

%\vspace{0.3cm}

\subsection{Accuracy of the learning model}
%\vspace{-2.5em} 
Once the deployment has been implemented on the edge device, a series of experiments were carried out to test how the prototype would work (according to the precision of the learning model) if it were to be deployed definitively in the factory.

In the experiments conducted, the model was trained using a labeled dataset consisting of $317$ samples collected from the cheese manufacturing production process (\Cref{data_desc}). To evaluate the accuracy of the model, we have used a set of $50$ production processes. 

When the percentage of predictions indicating defective products exceeds a specified threshold in the process, an alarm is triggered. This alert serves as a warning to the operator, indicating the need for immediate attention and optimization of the process to avoid further problems. ~\autoref{table:CM} shows the number of processes where the alarm was triggered and the number of processes where it really should have been triggered, i.e. when the product ended up being defective. The results are promising, 80.95\% of the times that alarms were triggered, the product was indeed defective, on the other hand, 75.86\% of the times that no alarms were triggered during the process the product was satisfactory.

\begin{table}
\begin{center}
\begin{tabular}{l*{2}{c}}
              & Alarm activated & Alarm not activated  \\
\hline
Defective process & 17 & 7  \\
Correct process            & 4 & 22 \\
\hline
\end{tabular}
  \vspace{0.2cm}
  \caption{Alarm confusion matrix}
  \label{table:CM}
\end{center}
\end{table}

These results suggest that the alarm system is effective in identifying potential defects, allowing operators to take timely corrective action. The accuracy rates indicate that the model can differentiate between satisfactory and unsatisfactory production results, minimizing the risk of defective products, thereby reducing waste and improving productivity.

\section{Conclusions}
\label{conclusions}

In this paper we have introduced an architectural proposal for IIoT settings where continual learning processes are really suitable, such as in the industrial field. Our modular architecture ensures scalability and easy deployment, allowing communication between its different modules. 

In order to validate our approach, we have selected a manufacturing company, Quescrem, that produces cheese to redefine its quality control process. The underlying idea was to provide a better solution able to detect defects in production in real-time.

The architecture was designed to be agnostic, so it can be applied to different factory environments beyond the initial use case. It also demonstrates the feasibility of implementing a continual learning model in an edge device within an industrial environment. This setup allows proactive intervention in the production process, improving product quality and operational efficiency. 

The results from the experiments conducted with data collected in the real factory are promising and show good performance. The continual learning model was effective in triggering alarms when product quality was at risk, with an accuracy rate of $80.95\%$ in predicting defective products when alarms were triggered and $75.86\%$ in predicting non-defective products when alarms were not triggered. These results suggest that the system has the potential for early detection of production problems, allowing operators to take corrective action in real-time.

Currently, we are working on the full deployment of the system in the manufacturing environment as well as in gathering data to improve the model. The next actions within the roadmap are: optimizing model handling instead of always using the latest trained model; developing a user interface to easily interact with the system; and scaling the implementation to allow the management and monitoring of multiple production lines simultaneously.

\section{Acknowledgement}{
This work was supported by the grant PID2020-113795RB-C33 funded by MICIU/AEI/10.13039/501100011033
(COMPROMISE project); the grant PID2023-148716OB-C31 funded by MCIU/AEI/10.13039/501100011033 (DISCOVERY project). The grant 101092069 funded by HORIZON-CL4-2022-TWIN-TRANSITION-01 (AI REDGIO 5.0 project) and it also has been funded by the Galician Regional Government under project ED431B 2024/41
(GPC).}

%\paragraph{Notes and Comments.}

%
% ---- Bibliography ----
%
\citestyle{acmauthoryear}
{\small
\nocite{*}
\bibliographystyle{unsrtnat}
\bibliography{myreferences}
}
\end{document}